# Implementation of an Onboard Visual Tracking System with Small Unmanned Aerial Vehicle (UAV)

Ashraf Qadir[1], William Semke[2], Jeremiah Neubert[3]
*Mechanical Engineering Department, University of North Dakota*
*Grand Forks, North Dakota-58201, USA*
[1] ashraf.qadir@my.und.edu
[2] williamsemke@mail.und.nodak.edu
[3] jeremiah.neubert@und.edu

*Abstract*—This paper presents a visual tracking system that is capable of running real time on-board a small UAV (Unmanned Aerial Vehicle). The tracking system is computationally efficient and invariant to lighting changes and rotation of the object or the camera. Detection and tracking is autonomously carried out on the payload computer and there are two different methods for creation of the image patches. The first method starts detecting and tracking using a stored image patch created prior to flight with previous flight data. The second method allows the operator on the ground to select the interest object for the UAV to track. The tracking system is capable of re-detecting the object of interest in the events of tracking failure. Performance of the tracking system was verified both in the lab and during actual flights of the UAV. Results show that the system can run on-board and track a diverse set of objects in real time.

I. INTRODUCTION

Visual tracking has been an active research topic due to its potential in wide range of applications in robotics and autonomous systems like vision based control [1], surveillance [2], augmented reality [3], visual reconstruction etc. One major research area is visual tracking with small UAV and its applications. Vision based tracking system has applications in vision based navigation [4], sense and avoid system [5], traffic monitoring [6], search and rescue, etc. They are particularly effective in GPS denied environment and tracking non-cooperative targets. For example, vision based tracking system can be used in tracking a car in an urban environment or rescue a person in woods.

Developments in auto pilot technologies and reduced cost has increased the potential applications of small unmanned aircraft systems. A considerable amount of work has been done on visual tracking with UAV and its application in vision based navigation, autonomous control and sense and avoid system for UAV [7, 8]. However, the major challenge in visual tracking with small UAVs is the computation cost of the tracking algorithm. Small UAVs have limited payload capacity and majority of the visual tracking systems are computationally expensive [11]. As a result most of tracking systems use powerful ground stations to run the tracking algorithm and the target position information is then sent to the UAV. The tracking system described in [9] uses an on-board camera to capture the video and sends the data to a ground computer that runs off-the-shelf (COTS) image processing software. The target information is extracted and guidance commands are then sent back to the UAV. These methods depend heavily on the communication between the UAV and the ground computer, thus a communication loss between the UAV and ground computer results in tracking failure. A linear parametrically varying (LPV) filter based motion estimation algorithm in their tracking system was proposed in [10] where the target-loss events due to communication interruption have been modeled as brief instabilities. However the algorithm shows a degradation of performance in presence of target loss events.

An on-board visual tracking system eliminates the dependency on the communication with the ground station and makes the system less prone to failure. However small unmanned aircraft systems have limited payload capacity and power budget. As a result an onboard visual tracking system requires a tracking algorithm which is robust and computationally efficient. An on-board visual tracking system for UAV control was described in [11]. The tracking system used a scale invariant feature transform (SIFT) algorithm for detecting salient points at every processed frame for visual referencing. Test results show satisfactory matching but at a rate not sufficient for real time tracking and they found tracking speed depends heavily on the size of the search window.

This paper presents a real-time visual tracking system developed to run onboard a small UAV. The algorithm is based on a similar system presented in [12]. Testing of the system was limited to simulated flight data. Based on the initial results, the system was changed to reduce the computational expense of the algorithm and improve its robustness. This improved system was evaluated using laboratory experiments and actual flight tests. The results show that the system is robust and capable of real-time operation on a small UAV.

The remainder of the paper is organized as follows: The basic architecture of the tracking system is described in Section (2). Section (3) describes the hardware and software implementation of the tracking system. Testing and results of the system are presented in Section (4). Future works with concluding remarks are presented in Section (5).

## II. METHODOLOGY

This section presents our tracking system, which includes: object detection using template matching, image warping, kalman filtering and camera actuation.

A Zero Mean Normalized Cross Correlation (ZMNCC) based template matching method was used for object detection as the object appears small from the UAV flying at 600 feet and its scale does not change significantly. Rotation invariance was achieved by creating a set of templates with 10 degrees interval using image warping and comparing the templates with the image frames. A kalman filter [13, 14] was used to make the system computationally efficient by limiting the search region in the image frame.

### A. Template Matching With Zero Mean Normalized Cross Correlation

A large number of tracking methods have been proposed for visual tracking. These methods vary in object representation and detection, and choice of a particular method depends on the application. A comprehensive description of different approaches for object representation and detection can be found in [15]. A vehicle on the ground from the UAV flying around 600 feet appears very small with the resolution as small as 20x20 pixels. Therefore it is difficult to extract enough features for feature based object detection. On the other hand, template matching techniques have been proved to be effective for recognition and classifying small objects.

The tracking system uses zero mean normalized cross correlation method to detect the object of interest in the video frames captured by the payload on-board the UAV. A small image patch of the interest object is used as the template and the object is detected comparing the template and the image using zero mean normalized cross correlation. The template is slid over the image and correlation coefficient is calculated to detect the position of the template in the image frame. A detailed description of normalized form of cross correlation can be found in [16, 17] where they also proposed fast algorithms to calculate zero mean normalized cross correlation coefficient. The Fast Normalized Cross Correlation Coefficient equation is described as

$$C = \frac{\sum [f(x,y) - \bar{f}_{u,v}][t(x-u, y-v) - \bar{t}]}{\left\{\sum [f(x,y) - \bar{f}_{u,v}]^2 [t(x-u, y-v) - \bar{t}]^2\right\}^{1/2}}$$

(1)

where $C$ is the Zero Mean Normalized Cross Correlation coefficient, $\bar{t}$ is the mean intensity value of the template and $\bar{f}_{u,v}$ is the mean intensity value of the image $f(x,y)$ in the region under the template. Subtracting the mean and normalizing the image and template make the correlation coefficient value ranges from -1 to +1. A best match between the image region and the template results in a coefficient value of +1 and -1 means a complete disagreement between the template and image. Different threshold values for the correlation coefficient ($C$) have been used in the algorithm during Experimentation with the previous flight videos. Results show that a correlation coefficient value of 0.9 or higher gives a true match between the image and template.

### B. Image Warping

Zero mean normalized cross correlation (ZMNCC) makes the tracking system invariant to intensity changes in the image sequences. However, template matching with ZMNCC detects object where there is only translation or small changes of the interest object shape or orientation. An object viewed from the UAV flying 600/700 feet appears small and its shape does not change much. But both the object and/or camera have rotation when tracking with a UAV. Image warping has been used to make the tracking system rotation invariant. A set of 36 templates have been generated from the original image patch with 10 degrees interval using image warping to accommodate full 360 degree rotation of the object. Templates are then compared with the image to detect the object of interest.

Image warping can be defined as mapping a position (x, y) in the source image to the position ($x'$, $y'$) in the destination image [18]. If a position in the source 2D image expressed in homogeneous coordinates as $x = [x, y, 1]^T$, and its corresponding position in the destination image as $x' = [x', y', 1]^T$ in homogeneous coordinates, then the mapping can be described as

$$x' = Hx \qquad (2)$$

where H denotes the Transformation matrix. For pure rotation the transformation is expressed as

$$\begin{bmatrix} x' \\ y' \\ 1 \end{bmatrix} = \begin{bmatrix} \cos\alpha & \sin\alpha & 0 \\ -\sin\alpha & \cos\alpha & 0 \\ 0 & 0 & 1 \end{bmatrix} \begin{bmatrix} x \\ y \\ 1 \end{bmatrix}$$

(3)

However not all 36 image templates are compared with the image to find the match every frame. The algorithm starts with comparing the first frame image with the templates. When a match is found, the algorithm leaves the cross correlation part of the program and the template number is recorded. In the next frame, the algorithm starts two patches previous to the patch number that was recorded in the previous iteration. However, the maximum number of templates compared in one frame is 7. If the algorithm does not find a match while comparing all 7 patches, it goes to the next frame. Now the algorithm starts at one patch after the one it started at in the previous frame. For example, once the system has the template, it generates 36 image templates with 10 degrees rotation. The algorithm starts cross correlation with template number 1 in the first frame. Say it finds the match at template number 4. The algorithm leaves the cross correlation and goes to the next frame. In the next frame, the algorithm will start with template number 2 and compare up to frame number 8. If there is no match in this frame, the algorithm goes to the next frame and starts the template matching with patch number 3.

*C. Kalman Filtering*

An extended Kalman filter has been used to make the tracking system computationally efficient and capable of running real time on-board the UAV. Moving the template over the entire source image and compute the correlation coefficient at every position is computationally expensive and consumes a lot of time. Predicting the position of the interest object in the next frame and searching only a region around the predicted position make the system computationally efficient. The position of the interest object is predicted using the motion model of the Kalman filter. Then a search window is generated around the predicted position using the process covariance matrix of the Kalman filter. The position is then estimated using the measurement from the object detection and the process covariance matrix is updated. In the absence of any detection, the state is not updated, another prediction is made and the detection process repeated. The covariance matrix gets bigger which in turn makes the search window bigger. Two governing equations for extended Kalman filtering are:

$$x_k = f(x_{k-1}, u_{k-1}, w_{k-1}) \qquad (4)$$
$$z_k = h(x_k, v_k) \qquad (5)$$

Equation (4) is the non-linear stochastic difference equation where $x_k$ is the state of the system, k is the time stamp, the function f describes the process model of the system and w is the process noise. Equation (5) relates the measurements $z_k$ to the state $x_k$ of the system at the time stamp k.

An extended Kalman filter is used for the vision tracking of the complex trajectory (change of acceleration) of the object [19]. Assuming a fixed velocity model, the state of the system is described as

$$X_k = \begin{bmatrix} \vec{x}_k \\ \dot{\vec{x}}_k \end{bmatrix}, \qquad (6)$$

where $X_k$ is the system's state, $\vec{x}_k$ is the object's position and $\dot{\vec{x}}$ is the velocity of the object at time instant $k$

The dynamic system was modeled as:
$$\vec{X}_k = \exp[\dot{\vec{x}}_{k-1}\Delta t] * \vec{X}_{k-1} \qquad (7)$$

The filter is initialized with the following items

Process Jacobian: For a 2 degree of freedom system the process Jacobian

$$A = \begin{bmatrix} 1 & 0 & \Delta t & 0 \\ 0 & 1 & 0 & \Delta t \\ 0 & 0 & 1 & 0 \\ 0 & 0 & 0 & 1 \end{bmatrix} \qquad (8)$$

Jacobian matrix $A$ is dependent on the time elapsed between observation k-1 & k and is denoted as $\Delta t$.

Process noise covariance matrix is initialized as

$$Q = \begin{bmatrix} a_1 & 0 & b_3 & 0 \\ 0 & a_2 & 0 & b_4 \\ b_3 & 0 & \Delta t\sigma_3 & 0 \\ 0 & b_4 & 0 & \Delta t\sigma_4 \end{bmatrix} \qquad (9)$$

where $\quad a_i = \Delta t\sigma_i + \frac{1}{3}\Delta t^3 \sigma_{i+2},$ (10)

and

$$b_i = \frac{1}{2}\Delta t^2 \sigma_i \qquad (11)$$

The values of $\sigma$ are determined using experimentation with the previous flight data. The search windows were generated from the covariance matrices around the predicted position. A value of 0.4 for $\sigma$ generates a search window big enough to keep the object inside the window.

The Measurement Jacobian is

$$H_k = \frac{\partial h}{\partial \overline{X}} = \begin{bmatrix} 1 & 0 & 0 & 0 \\ 0 & 1 & 0 & 0 \end{bmatrix}, \qquad (12)$$

Measurement noise $R_k$ is projected into state space using the equation $V_k R_k V_k^T$. The following measurement noise matrix here has been used: $R_k = \begin{bmatrix} 1 & 0 \\ 0 & 1 \end{bmatrix}$,

(13)

where the elements represents one pixel of uncertainty in object localization in the image frame. The steps involved in using Kalman filtering in our vision tracking system are:

1. Initialization ($k$=0): In this stage the whole image is searched for the object due we do not know previously the object position. The object is detected in the image frame and its centre is selected as the initial state $\hat{x}_0$ at time $k=0$. Process covariance matrix $\hat{P}_k$ is also initialized.

2. Prediction (k>0): The state of the object $\hat{x}_{k=1}^-$ is predicted using the motion model of the Kalman filter for the next image frame at time k=1. This position is considered as the center of the search window to find the object.

3. Correction (k>0): In this stage the object is detected within the search window (measurement $z_k$) and the state $\hat{x}_k$ and covariance matrix $P_k$ is updated with the measurement data.

Steps 2 and 3 are carried out while the object tracking runs. The size of the search window is dictated by the noise in the prediction and depends on the process covariance matrix. A small process covariance matrix results in a small search window size and implies that the estimation is trusted more. In the absence of detection, there are no measurements and state is not updated. This results in a larger process error covariance and the search window gets bigger.

*D. Actuating the Gimbal*

Once detected, the distance between the center of the image and the detected position is computed. The distance is then converted to motor count and sent to the motion controller for the pan and tilt motion of the gimbal. The gimbal actuates accordingly to keep the detected object at the center of image frame.

III. IMPLEMENTATION

The tracking system was implemented on the SUNDOG (Surveillance by University of North Dakota Observational Gimbal) payload [19, 20] developed by the undergraduate mechanical and electrical students at the University of North Dakota. A customized UAV named "Super Hauler", owned and operated by the UASE lab, UND houses the SUNDOG payload as well as the Piccolo autopilot [21] with its dedicated control link. The UAV has a wingspan of 144 inches and 120 inches of length. The UAV weighs 48 pounds and has

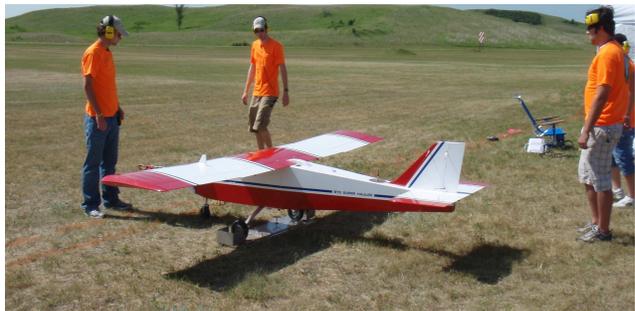

25 pounds of payload carrying capacity.

Fig. 1. BTE Super Hauler UAV

*A. SUNDOG Payload*

The payload consists of a PC/104+ form factor based computer- essentially a Linux PC on a single printed circuit board (PCB) with frame grabber, additional octal serial port board and wireless card, a three-axis precision pointing system for an Electro-Optical camera and an Infrared camera. A 2.4 GHz PC/104-plus form factor based wireless card and a RTD PC/104- Plus Dual Channel Frame Grabber is stacked with the computer. A color Sony FCBEX980 camera is mounted on the gimbal. The gimbal has 360 degrees rotation and 30 degrees pan and tilt rotation motion.

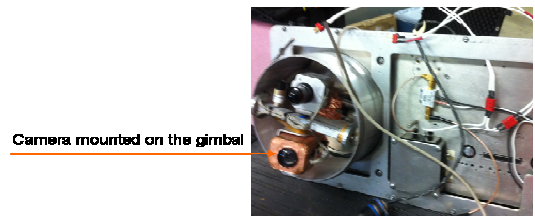

Fig. 2. SUNDOG [19, 20] payload. Image shows the three-axis gimbal system with the Electro-Optical (EO) and Infrared (IR) cameras mounted on it.

*B. Motion Controller*

FaulHaber motion controller [22] use pulse-width modulation (PWM) signals to drive the DC servo motors. The drive or amplifier transforms the PWM signal into high amplitude current to turn the motors. They allow for torque control via current regulation. Incremental encoders have been used for position feedback. The controllers are connected via RS232 serial cable to the on-board computer and provide resolution of 100 micro-radians (0.00570).

*C. Joystick Control*

A joystick control of pointing the gimbal has been implemented for on-line selection of the interest object from the ground. Joystick control allows the ground operator to manually point the gimbal at the target for selection.

IV. EXPERIMENTATION AND RESULTS

*A. Testing with Previous Flight Video*

The algorithm was tested with video captured in previous test flights with two objectives in mind. One is to check that our detection algorithm is robust enough to locate objects on the ground and the second one is to tune the algorithm so that it is computationally efficient and capable of real time tracking. The algorithm was tested with different threshold values for the normalized cross correlation coefficient and search window size. The search window size depends on the covariance matrix of the Kalman filter. Process covariance matrices were generated by selecting different values of covariance matrix elements ($\sigma$). Then the system was optimized by generating the search window which is big enough to make sure that the object stays inside the search window in successive frames. An object is selected in one image frame and then tracked in subsequent frames, as shown in the Figure 3. The object is selected in the left image and the template is generated. The red rectangle represents the selection area in the source image frame. The right image shows the object tracking. The red dot in the image represents the detected object which is the centroid of all matching points between the template and the source image. The blue rectangle around the object is the search window generated by the covariance matrix of the Kalman filter while the red rectangle represents the initial position of the object.

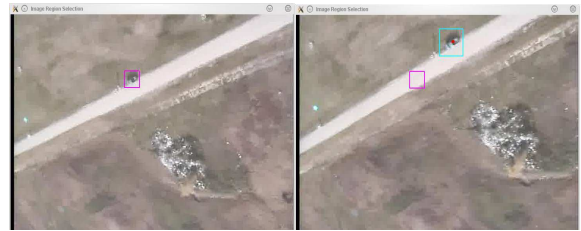

Fig. 3. Tracking on the previously captured image frames. The object was selected on the left frame and the right frame shows the object detection.

The recorded data was used to determine the optional parameter settings for in-flight operation. The initial testing was used to select the threshold for ZMNCC. This was accomplished by tracking several objects in the video sequence with differing threshold values. The selected value, 0.9, was able to successfully match the desired object more than 95% of the time with no false positives.

After the threshold was selected, the video was used to determine the noise in our process model. This was accomplished by assuming a small amount of uncertainty and attempting tracking. The tracking algorithm was limited to a search region based on predicted uncertainty assuming that the object remains within three standard deviations of its predicted region. If an interest object fell outside the search region, the estimated model uncertainty was increased. The process continued until the uncertainty allowed all the interest objects in the video sequence to be maintained in the predicted search region. Using a model uncertainty matrix with 0.4 on the diagonal produced the desired results.

*B. Tracking with Model Cars*

A laboratory experiment has been carried out by simulating the flight environment. The objective of the experiment was to verify that the tracking algorithm was robust and computationally efficient. The algorithm was also tuned with the experimental data. A small moving car was used as the target object. The payload was mounted on a stand 3 feet (0.9144 meters) above the ground. The goal of the payload was to track the car on the ground and move its gimbal to keep the car at the center of the image frame. The middle section of the top of the car was selected as the template. The tracking algorithm detected the car and started tracking. To verify the robustness and efficiency of the tracking system, the car was translated, rotated and moved around other objects. Tracking was displayed on the screen in real time as the system was running and intermittent image frames were saved for later analysis purpose.

Online selection of an object of interest and sending this data to the payload was also tested in the lab using a direct serial communication between ground computer and the payload. Video frames were sent over the Ethernet from the payload to the computer used as a ground station. The car was selected on the ground computer screen and the image patch was sent to the payload using the RS232 communication. Upon receiving the image patch, the payload started tracking.

Results of the experiment are shown in Figure 4. The red dot shows the centroid of the detected positions and the green rectangle around it is the search window generated by the covariance matrix of the Kalman filter. The object appeared with different position and orientation as it was moved and the tracking system was able to detect the car.

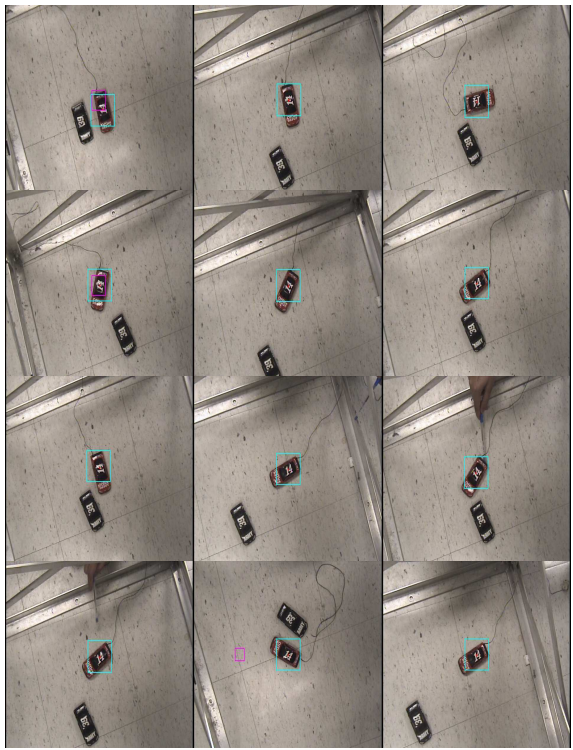

Fig. 4. Tracking model car in the lab. The car was translated and rotated to test the performance of the tracking algorithm

The gimbal was actuated to keep the car at the center of the image frame. The tracking system exhibited its ability to real time track and re-detect objects in the event of tracking failure in one frame. The system was able to predict the state of the tracked object and the search region was able to keep the object inside.

Computation cost of the tracking system was also verified by computing the frame rate the tracking system was capable of processing. Results show that the system was able to track real time with more than 25 frames per second. Results also show that the tracking speed varies with the patch sizes. Different size patches were selected during the tests and being tracked by the algorithm. Smaller patches resulted in higher tracking speed. Table 1 shows the frame rates for different size patches.

TABLE 1

TRACKING SPEED WITH DIFFERENT PATCH SIZES

| Patch Size (Pixels) | Number of Frames | Tracking Speed (Frames/Sec) |
|---|---|---|
| 27x28(756) | 813 | 27.83 |
| 20x22(440) | 246 | 28.21 |
| 38x30(1140) | 558 | 26.99 |
| 30x33(990) | 533 | 27.57 |

### C. Actual Flight Tests

Finally multiple flight tests of the complete tracking system were conducted in May-August of 2011. The flight tests validate the tracking algorithm and shows that the tracking system is capable of re-detecting the interest object in the event of tracking failure. The test bed for the tracking system is explained in Fig. 5.

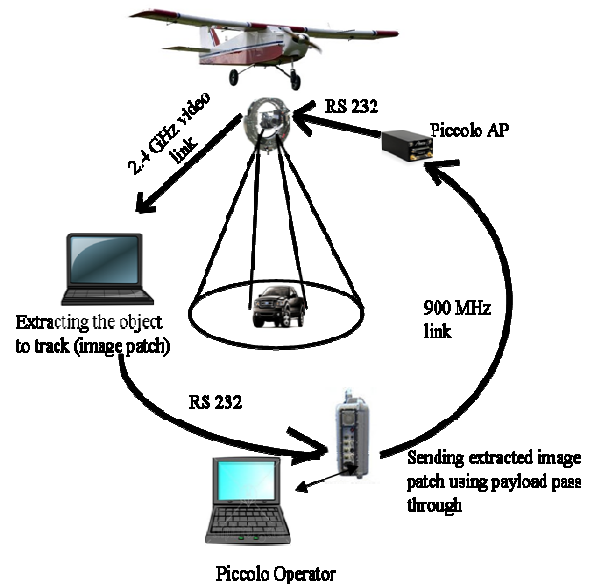

Fig. 5. Flight test bed

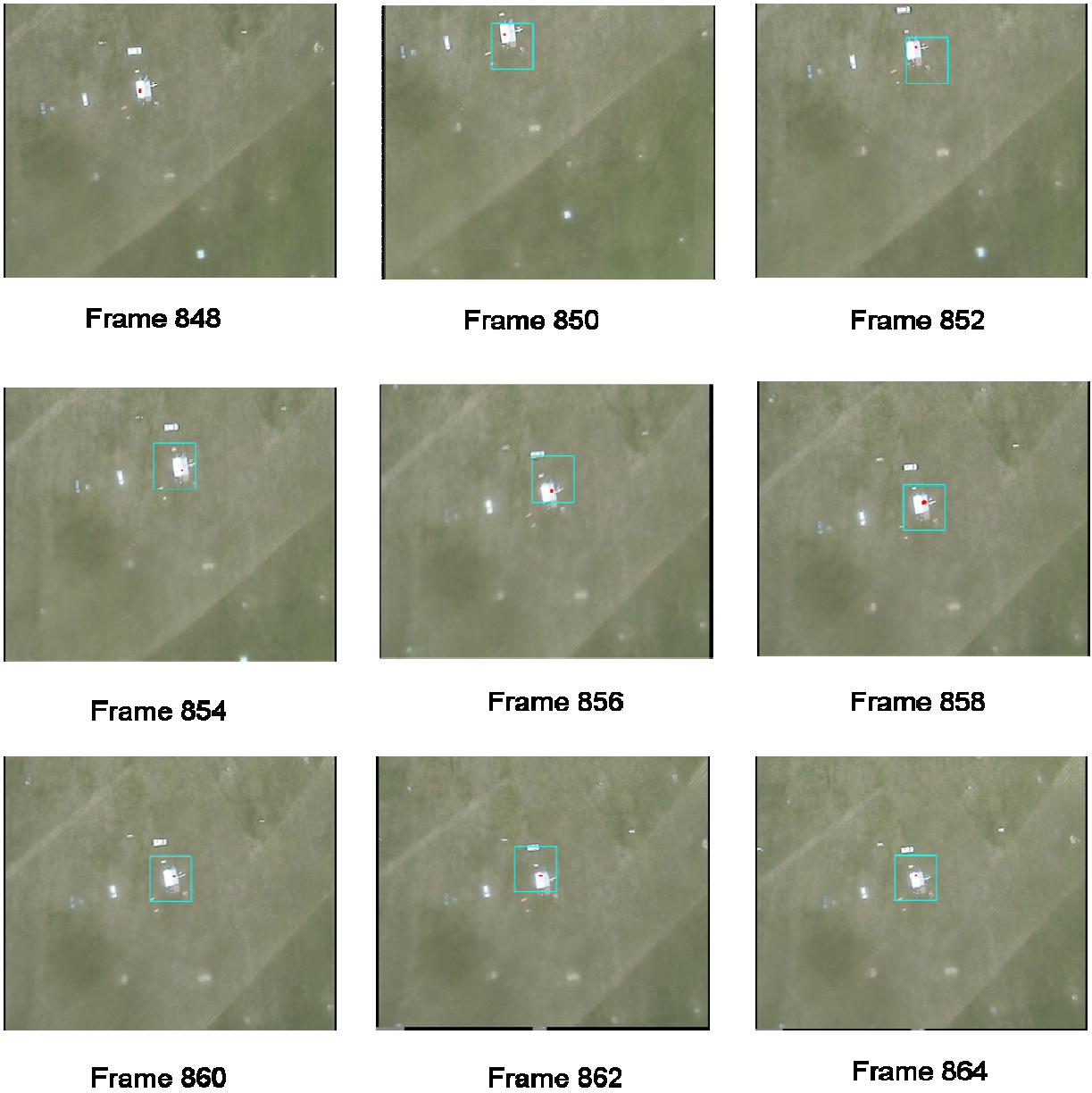

Fig. 6. Tracking results from actual flight test. The patch is shown at the bottom left corner of the first image frame. The object was detected in one video frame. The camera was then actuated to keep the object at the centre of the image frame. The red dot shows the detection of the object and the blue rectangle is the search region around the object. In the first frame there is no blue rectangle drawn because it is the first frame where the object was detected and the search region was the whole image. Once the object was detected, the covariance matrix of the Kalman filter was updated with the measured data and the search region gets smaller.

The UAV was flying at 600 feet altitude and a repeated flight path over the ground control tent was chosen for the UAV. The video captured by the payload camera was broadcast to the ground computer using 2.4 GHz communication link. On the ground the interest object was selected from the video and sent to the UAV during one pass of the UAV over the object. Once the image patch is received by the payload, the algorithm generates 36 image patches of the interest object with 10 degrees rotation to cover the whole 360 degrees rotation of the object. The object was lost when it went out of the camera view. Once the UAV came over the object again in the next pass, it detected the object and started tracking. Then the gimbal with the camera was moved to keep the object at the center of the image frame. The tracking system successfully tracked four different objects during the flight tests. The results of the flight test have been shown in Fig. 6.

## V. CONCLUSION

An on-board visual tracking with small UAV that is capable of tracking real time has been developed. The tracking system locates the object of interest using zero mean normalized cross correlation between object template and source image. The tracking system is invariant to changes in illumination or rotation of the object. The tracking system uses a Kalman filter to estimate the object position and create a search window around the estimated position. Image warping has been used to make the tracking system rotation invariant. The system was implemented on the SUNDOG payload and was tested using several experimentations including a full hardware in the loop test in the lab and actual flight tests. The tracking algorithm has the ability to re-detect and track in the event of loss of tracking. If the object goes out side the search window the tracking fails. The system quickly recovered from this failure by expanding the search area. The system assumes that large scale changes of the tracked object do not occur. This assumption appears valid during actual flight tests. A simple linear projective transformation can be used in the tracking system with the aircraft altitude information to accommodate scaling. Better control of the camera parameters such as gain, and exposure will reduce the blurriness in the images caused by vibration or very fast movement of both the camera and the object.

## ACKNOWLEDGMENT

This research was supported in part by Department of Defense contract number FA4861-06-C-C006, "Unmanned Aerial System Remote Sense and Avoid System and Advanced Payload Analysis and Investigation," and North Dakota Department of Commerce grant entitled "UND Center of Excellence for UAV and Simulation Applications." The authors appreciate the contributions of the Unmanned Aircraft Systems Engineering (UASE) Laboratory team at the University of North Dakota.